\documentclass{amia}
\usepackage{float}
\usepackage{amsmath}
\usepackage{booktabs}
\usepackage{enumitem}
\makeatletter
\renewcommand\@cite[2]{\textsuperscript{#1}}
\makeatother

\begin{document}

\title{Machine Learning for Pre-Culture ESBL Risk Stratification to Guide Empiric Antibiotic Selection: A 12-Hospital Study of Enterobacteriaceae Cultures}

\author{
Aravind V. Kuruvikkattil, BAMS, MS$^{1}$,
Lalitha Pranathi Pulavarthy, BDS, MS$^{1}$,
Rashmita Kudamala, BPT, MS$^{1}$,
Saptarshi Purkayastha, PhD$^{1}$
}

\institutes{
$^{1}$ Dept. of Biomedical Engineering and Informatics, Luddy School of Informatics, Computing, and Engineering,
Indiana University Indianapolis, IN
}

\maketitle

\section*{Abstract}

\textit{%
Empiric antibiotic therapy for suspected ESBL-producing Enterobacteriaceae must be selected 48-72 hours before culture results, forcing clinicians to choose between undertreating resistant infections and overusing carbapenems that drive further resistance. We developed a cost-sensitive XGBoost model predicting an ESBL phenotype (resistance to ceftriaxone, ceftazidime, cefepime or piperacillin-tazobactam) at culture ordering using 45 pre-culture EHR features across 132,955 cultures from 72,217 patients at 12 hospitals (14.41\% with the ESBL phenotype). Cultures were partitioned at the patient level. At 90\% sensitivity, the model achieved 95.8\% NPV, reducing post-test ESBL probability to 4.2\%, a threshold that may support safe carbapenem-sparing in non-ICU settings, while sparing 307 of every 1,000 cultures an unnecessary broad-spectrum course at the cost of 14 missed ESBL cases per 1,000. SHAP analysis identified prior ESBL colonization as the dominant predictor, ahead of prior organism burden and neighborhood deprivation; removing deprivation features caused minimal performance loss
($\Delta$AUROC\,=\,$-$0.020), enabling equitable bedside deployment. Discrimination was unchanged under a strict IDSA ESBL-E definition (AUROC 0.766), with specimen type added as a predictor (0.764) and without any class-imbalance correction (0.762), and ranged from 0.71 to 0.78 across organism strata.%
}

\section*{Introduction}

Extended-spectrum beta-lactamase (ESBL)-producing Enterobacterales are among the fastest-growing antimicrobial resistance threats: in the United States, hospital-associated infections rose by about 50\% between 2013 and 2017 to an estimated 197,400 cases annually~\cite{cdc2019}. These infections usually require carbapenems, and delayed or inadequate escalation increases 30-day mortality, length of stay, and cost~\cite{schwaber2007,tumbarello2007}.

The fundamental clinical challenge is temporal: empiric antibiotic therapy must be prescribed immediately upon culture collection, while susceptibility results are unavailable for 48-72 hours. During this window, clinicians must balance two competing risks: prescribing narrow-spectrum agents that may fail against ESBL-producing organisms, or prescribing carbapenems unnecessarily and contributing to the selection pressure driving carbapenem-resistant Enterobacterales (CRE), which carry mortality rates exceeding 40-50\%~\cite{gupta2011}. Because carbapenems remain the recommended therapy for serious ESBL infections, and piperacillin-tazobactam proved inferior for ESBL bacteraemia in the MERINO trial~\cite{tamma2024,harris2018}, the stewardship opportunity lies in confidently identifying the majority of patients whose isolates are unlikely to be ESBL-producing and who can safely receive narrower empiric therapy.

Machine learning models predicting antibiotic resistance at the time of culture ordering have emerged as a promising decision-support strategy~\cite{kanjilal2020,yelin2019,corbin2022,goto2026}. Structured EHR features predict extended-spectrum cephalosporin resistance or ESBL production with moderate discrimination: gradient boosting reached AUROC 0.74 for ceftriaxone resistance in Enterobacterales bacteraemia~\cite{yuan2025}, 0.80 in an emergency-department ESBL model~\cite{kuzmich2025}, and 0.78--0.81 for extended-spectrum cephalosporin resistance in a multitask model across the Veterans Health Administration~\cite{goto2026}. Across these studies, personal culture history is consistently the strongest predictor~\cite{yelin2019,macfadden2018}. Separately, ecological and cohort studies link neighborhood deprivation to higher resistance prevalence and worse outcomes in Enterobacterales infections~\cite{cooper2024,henderson2025}, and socioeconomic deprivation to elevated community antibiotic prescribing and AMR burden~\cite{dolk2018,collignon2018,raju2025,kuruvikkattil2026}, but the Area Deprivation Index (ADI)~\cite{kind2018} has rarely been tested as a patient-level predictor alongside clinical history.

However, gaps remain. Most ESBL models come from single health systems and target bloodstream infections exclusively~\cite{goodman2016,tumbarello2011,rottier2018,yuan2025}; the relative weight of neighborhood deprivation against clinical history has not been quantified within one model; and few studies report whether validation partitions were constructed at the patient level, even though repeat cultures make record-wise splits optimistic~\cite{saeb2017,kapoor2023}.

We address these gaps using the ARMD-MGB dataset, a de-identified, multi-hospital EHR-linked microbiology repository containing 225,000+ patients across 12 Mass General Brigham hospitals~\cite{PhysioNet-armd-mgb-1.0.0}. We developed and validated an XGBoost model predicting an ESBL phenotype at the moment of culture ordering across 132,955 Enterobacteriaceae cultures spanning all culture types and care settings. This study had three objectives: (1)~\textit{develop} a clinically actionable prediction model whose negative predictive value is sufficient to support safe empiric carbapenem-sparing; (2)~\textit{quantify}, via SHAP attribution, the relative contribution of prior colonization history and neighborhood deprivation against other clinical risk factors; and (3)~\textit{demonstrate} leakage-free performance under patient-level partitioning, organism-stratified stability, and clinical utility through decision curve analysis.

\section*{Methods}
\subsection*{Dataset and Study Population}
We used the Antibiotic Resistance Microbiology Dataset from Mass General Brigham (ARMD-MGB), a de-identified, multi-site EHR-linked microbiology repository available through PhysioNet~\cite{PhysioNet-armd-mgb-1.0.0}, the Mass General Brigham counterpart of the Stanford ARMD resource~\cite{nateghi2025}. The dataset encompasses 225,000+ patients across 12 MGB-affiliated hospitals from 2015 through 2024. The study unit was a single culture order. We included all Enterobacteriaceae cultures with a documented organism, antimicrobial susceptibility testing (AST), and complete metadata, yielding 132,955 analyzable cultures from 72,217 unique patients (mean 1.84 cultures per patient; 64.2\% of cultures came from patients contributing more than one culture). Because the de-identified ARMD-MGB release does not expose a stable site identifier, patient-mix differences across the 12 contributing hospitals could not be directly characterized or adjusted for at the culture level; robustness to between-population variation is instead assessed indirectly through organism-stratified and subgroup analyses below.

\subsection*{Outcome Definition}
\paragraph{Primary definition (ESBL phenotype).}
The primary outcome is an ESBL phenotype defined from CLSI 2022 interpretations: a culture was labeled positive if any of four sentinel beta-lactam agents, ceftriaxone (CRO), ceftazidime (CAZ), cefepime (FEP), or piperacillin-tazobactam (TZP), was reported as resistant on AST. This yielded 19,159 positive cultures (14.41\%). The composite deliberately captures the resistance pattern that prompts empiric carbapenem use rather than a single enzyme class: it includes piperacillin-tazobactam resistance, which is not part of CLSI ESBL screening criteria, and cultures from genera with chromosomal AmpC beta-lactamases (\textit{Enterobacter}, \textit{Citrobacter}, \textit{Serratia}, \textit{Morganella}, \textit{Providencia}), in which ceftriaxone resistance is usually AmpC- rather than ESBL-mediated. Treatment decisions rest on the susceptibility phenotype rather than molecular characterization, and the definition is comparable to the extended-spectrum cephalosporin resistance outcomes used in prior prediction studies~\cite{rottier2018,yuan2025}. Because IDSA guidance applies the ESBL-E label to ceftriaxone-non-susceptible \textit{E.\ coli}, \textit{K.\ pneumoniae}, \textit{K.\ oxytoca} and \textit{P.\ mirabilis}~\cite{tamma2024}, we also report a sensitivity analysis restricted to those species with ceftriaxone resistance as the outcome.

\paragraph{Sensitivity analysis (enzyme-class).}
As secondary analysis, we evaluated model performance on \texttt{enzyme\_class} field available in ARMD-MGB, which documents the confirmed resistance mechanism at the isolate level (ESBL, carbapenemase, beta-lactamase, or mecA/PBP2a). This enzyme-level classification is methodologically distinct from the phenotype definition and may capture a narrower population of molecularly confirmed ESBL producers~\cite{canton2012}.

\subsection*{Feature Engineering}
We constructed 45 predictive features drawn exclusively from structured EHR data available \textit{at the time the culture order was electronically signed} ($t_0$). No data generated after $t_0$, including Gram stain results, preliminary susceptibilities, or culture growth notifications, were included in the feature set, ensuring strict
temporal separation between predictors and outcome. Features were organized into six domains:

\textit{(1) Demographics:} age (ordinal-encoded), sex, age $\geq$65 indicator.

\textit{(2) Socioeconomic:} Area Deprivation Index (ADI), a validated census-block-level composite measure of socioeconomic disadvantage ranging from 1 (least deprived) to 100 (most deprived)~\cite{kind2018}. Of 132,955 cultures, 42.6\% had missing ADI due to ungeocoded addresses; a binary missingness indicator was included as a separate feature, and missing ADI values were set to zero rather than imputed with the population median, to avoid creating an artificial data cluster.

\textit{(3) Comorbidities:} six binary Elixhauser conditions (heart failure, liver disease, lymphoma, metastatic cancer, obesity, renal failure) and total Elixhauser comorbidity count.

\textit{(4) Prior antibiotic exposure:} binary indicators for use of seven antibiotic classes in the 90 days preceding culture collection (fluoroquinolones, third-generation cephalosporins, carbapenems, glycopeptides, sulfonamides, extended-spectrum penicillins, aminoglycosides). The 90-day window was selected based on causal evidence that this horizon  captures the window of maximal selective pressure~\cite{costelloe2010,kuruvikkattil2026}.

\textit{(5) Prior resistant organisms:} binary flags for prior documented ESBL, CRE, MRSA, and VRE; total number of prior cultured organisms; and days elapsed since most recent prior culture.

\textit{(6) Procedures and care setting:} binary flags for central venous catheter, mechanical ventilation, hemodialysis, urinary catheterization, and recent surgical procedure; nursing home residence within 90 days; and inpatient, emergency department, outpatient, or urgent care ward classification.

Eight interaction features captured clinically relevant compound exposures (e.g., concurrent fluoroquinolone and cephalosporin use, prior ESBL with recent cephalosporin exposure).

\subsection*{Model Development}
\noindent\textbf{Patient-level data partitioning.}~Cultures are nested within patients, so a culture-level random split would place 59.8\% of test cultures in a patient also seen in training, letting the model memorise patient-specific labels~\cite{saeb2017,kapoor2023}. We therefore partitioned 80\%/20\% by stratified group $k$-fold on the de-identified patient identifier (\texttt{anon\_id}), keeping every patient wholly within the derivation set (106,586 cultures, 57,742 patients) or the held-out test set (26,369 cultures, 14,475 patients); patient overlap was verified zero. All inner cross-validation (tuning, calibration, stability) was likewise patient-grouped. The test set was withheld from all development decisions, and we repeated the comparison under the culture-level random split as a sensitivity analysis.

\noindent\textbf{Class imbalance.}~Class imbalance (14.41\% positive) was addressed with two strategies, evaluated head-to-head: (1)~Synthetic Minority Oversampling Technique (SMOTE, k\,=\,5)~\cite{chawla2002} applied to the training set, and (2)~cost-sensitive learning with \texttt{scale\_pos\_weight} set to the inverse class frequency ratio in the gradient boosting objective. Both strategies were applied to all three model families. Cost-sensitive learning without SMOTE yielded superior or equivalent discrimination across all models and was selected as the primary approach. Because both strategies inflate predicted probabilities~\cite{goorbergh2022}, all reported probabilities were subsequently recalibrated.

Three model families were trained: $\ell_2$-penalized logistic regression (C\,=\,1.0, SAGA solver), random forest (300 estimators, maximum depth 8, minimum samples per leaf 20), and XGBoost~\cite{chen2016} (500 estimators, maximum depth 6, learning rate 0.05, subsample 0.80, column subsample 0.80). All features were standardized prior to logistic regression training; tree-based models received raw values.

\noindent\textbf{Hyperparameter optimization.}~XGBoost hyperparameters were optimized using Optuna with TPE sampling over 50
trials~\cite{akiba2019}, each evaluated via 3-fold patient-grouped stratified cross-validation on the training set. The search space spanned max\_depth (3--10), learning rate (0.01--0.3), n\_estimators (100--1000), min\_child\_weight (1--10), subsample and colsample\_bytree (0.5--1.0), gamma (0--5), and L1/L2 regularization (0--10). The optimization objective was average precision (AUPRC). The best configuration was retrained on the full training set and evaluated on the held-out test set.

\noindent\textbf{Calibration.}~Cost-sensitive weighting inflates predicted probabilities, so a Platt scaling map was fitted on 5-fold patient-grouped out-of-fold predictions within the training set and applied to the final model's test-set scores; this leaves rankings, and therefore AUROC and AUPRC, unchanged and corrects only the probability scale~\cite{vancalster2015}. Platt scaling was pre-specified as the primary method, and isotonic regression fitted the same way is reported as a sensitivity check. The same procedure was applied to the comparator models so that decision curves and operating points share one probability scale. Calibration was assessed by the Brier score and by calibration intercept and slope before and after recalibration. Model stability was assessed with patient-grouped stratified 5-fold cross-validation.

\noindent\textbf{Order-time partition.}~We additionally retrained all three families on an 80/20 partition ordered by culture order time. ARMD-MGB shifts all dates randomly per patient while preserving within-patient consistency, so this ordering does not recover calendar time and the partition is not a temporal validation; we report it as a second, near-patient-disjoint partition and characterise its date-shift properties in the Results.

\noindent\textbf{Organism-stratified evaluation.}~Because ESBL prevalence and resistance patterns vary across genera, we evaluated the pooled model separately on three organism groups: \textit{Escherichia coli}, \textit{Klebsiella pneumoniae}, and other Enterobacteriaceae. AUROC and AUPRC were computed for each stratum to assess whether pooled training generalizes across species.

\subsection*{Evaluation}
Primary discrimination was assessed by AUROC on the held-out test set. Precision-recall performance was assessed by AUPRC, with a no-skill baseline equal to ESBL prevalence in the test partition (0.137). Calibration was assessed by the Brier score. Three clinical operating points were defined on the recalibrated probability scale: an \textit{alert} threshold targeting $\geq$90\% sensitivity, mirroring the $\geq$90\% susceptibility convention used in cumulative antibiograms to define acceptable empiric coverage~\cite{hindler2007}. We also set a \textit{balanced} threshold maximizing F$_1$ score, and a \textit{conservative} threshold targeting $\geq$80\% specificity. Sensitivity, specificity, PPV, and NPV were reported at each threshold. Bootstrap 95\% confidence intervals (2,000 iterations) were computed for all primary metrics by resampling whole patients rather than individual cultures, since with repeat cultures the effective sample size is the number of patients.

\noindent\textbf{Decision curve analysis.}~DCA was performed on recalibrated probabilities for all three models across threshold probabilities 0.01--0.80~\cite{vickers2006}, because net benefit is interpretable only when predicted probabilities are calibrated~\cite{vancalster2015}. Net benefit was computed as $\text{NB} = \frac{\text{TP}}{N} - \frac{\text{FP}}{N} \cdot \frac{p_t}{1 - p_t}$. XGBoost, logistic regression, random forest, treat-all, and treat-none strategies were compared.

\noindent\textbf{Statistical comparison.}~Pairwise AUROC differences between models were assessed using the DeLong test~\cite{delong1988}, which accounts for correlation between AUROCs computed on the same test set. Net Reclassification Improvement (NRI)~\cite{pencina2008} was computed on recalibrated probabilities comparing XGBoost versus logistic regression using four risk categories ($<$10\%, 10--30\%, 30--60\%, $>$60\%); category-based and continuous NRI are reported with 2,000-iteration bootstrap 95\% confidence intervals, together with their event and non-event components, because category NRI is sensitive to calibration and to the cut-points chosen.

\noindent\textbf{Interpretability and subgroup analysis.}~SHAP TreeExplainer values~\cite{lundberg2017} were computed on the tuned model over 2,000 random held-out test observations; feature importance was summarized as mean $|$SHAP$|$. Subgroup AUROC was computed across sex, ward type, ADI level, and prior resistance status. False negatives were compared to true positives using Mann-Whitney U-tests.

\noindent\textbf{Additional analyses.}~Five analyses address reviewer-level concerns. (1)~An XGBoost model with no imbalance correction (no weighting, no SMOTE) was trained with the default hyperparameters to test whether cost-sensitive weighting improves discrimination at all, as simulation work predicts it should not~\cite{goorbergh2022}. (2)~Because DeLong's test assumes independent observations and the test set holds 1.8 cultures per patient, AUROC differences between models were also given 95\% confidence intervals from the patient-clustered bootstrap. (3)~Because missing ADI is zero-filled, a value outside the 1--100 range that trees can exploit as a geocoding flag, the tuned model was refitted with ADI encoded as a native missing value while retaining the missingness indicator, SHAP importance was recomputed, and the ADI score was ablated alone with the indicator retained. (4)~Specimen type (urine, blood, respiratory), which is known at order time but absent from the 45 features, was evaluated both as a stratifier of the reported model and as three indicator features in an extended 48-feature model. (5)~The strict ESBL-E sensitivity analysis restricted the cohort to the four IDSA species with ceftriaxone testing, used ceftriaxone resistance as the label, and both scored the reported model against that label and retrained the tuned configuration on it with the same patient-grouped partition.

All analyses used Python 3.11 (scikit-learn 1.4, XGBoost 2.0, SHAP 0.43, Optuna 3.6) with GPU acceleration (NVIDIA RTX 3090).

\noindent\textbf{Code availability.}~The complete analysis pipeline, including the patient-level partitioning and every result reported here, is available at {\urlstyle{tt}\url{https://github.com/iupui-soic/AMIA_2026_PriorESBL}}. ARMD-MGB is credentialed-access under a PhysioNet data use agreement and is not redistributed.

\section*{Results}
\subsection*{Study Population}
\begin{figure}[!t]
\centering
\includegraphics[width=\columnwidth]{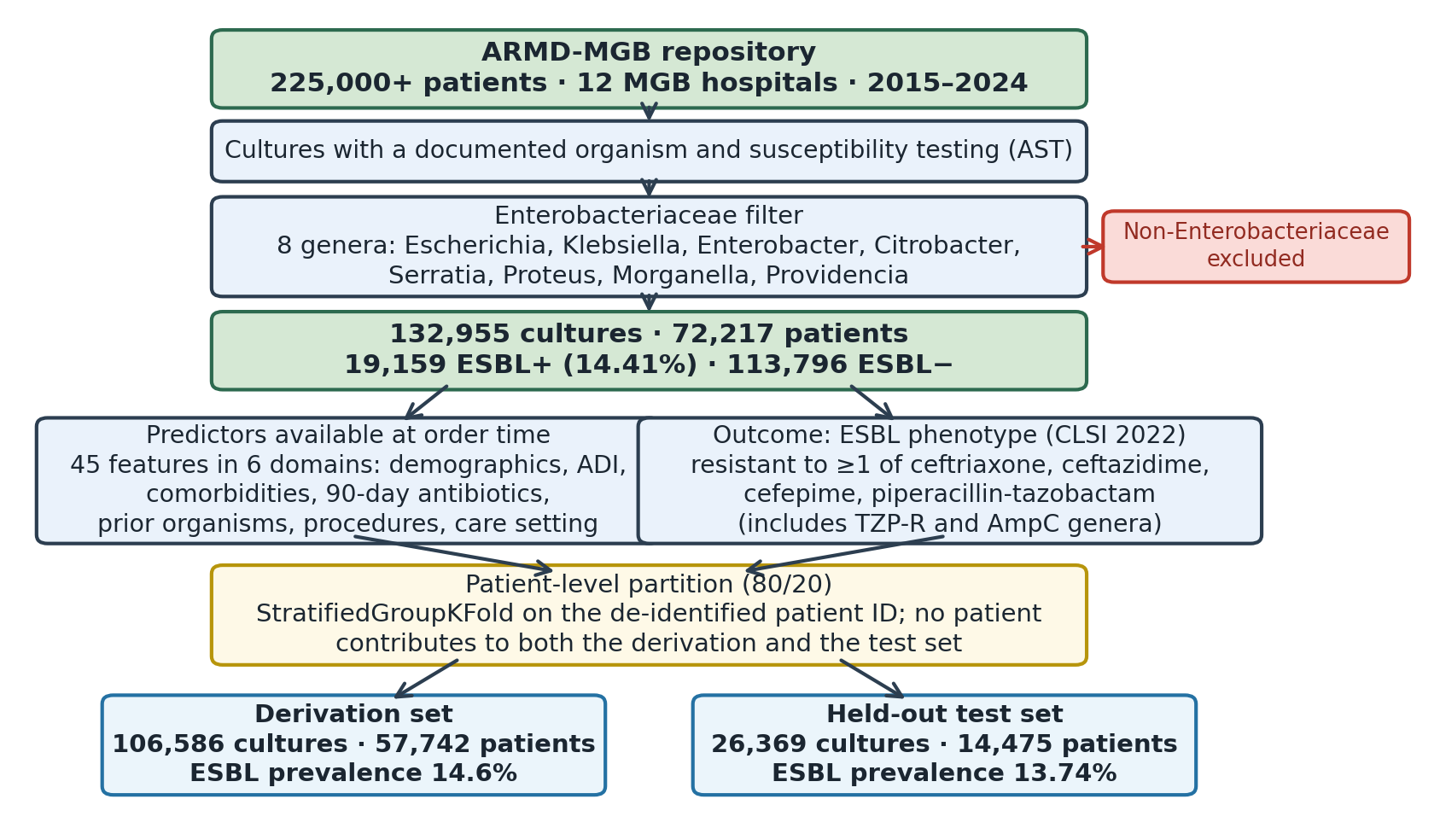}
\caption{CONSORT/TRIPOD-style cohort selection flowchart. Starting from the ARMD-MGB repository (225,000+ patients, 12 hospitals, 2015-2024), cultures were filtered to Enterobacteriaceae with documented organism and AST results, yielding 132,955 analyzable cultures (14.41\% ESBL-phenotype positive), partitioned at the patient level into derivation and held-out test sets.}
\label{fig:flowchart}
\end{figure}

Of 132,955 Enterobacteriaceae cultures from 72,217 patients, 19,159 (14.41\%) met the ESBL-phenotype definition. The most frequent organisms were \textit{Escherichia coli} and \textit{Klebsiella pneumoniae}. Cultures originated from inpatient, emergency department, outpatient, and urgent care settings across all 12 MGB hospitals. ADI data were available for 57.4\% of cultures; missing ADI was handled via a binary indicator feature with zero imputation. Table~\ref{tab:cohort} summarizes cohort characteristics overall and by ESBL-phenotype status. Phenotype-positive cultures were enriched for prior resistant-organism history (prior ESBL 40.4\% vs.\ 12.5\%; SMD\,+0.67), recent broad-spectrum exposure (90-day carbapenem 16.3\% vs.\ 2.6\%; third-generation cephalosporin 43.8\% vs.\ 25.4\%), and inpatient origin (47.7\% vs.\ 28.1\%), whereas demographic and neighborhood-deprivation distributions were similar across groups (ADI SMD\,+0.01). Figure~\ref{fig:flowchart} summarizes the cohort selection process, reported in line with TRIPOD+AI~\cite{collins2024}.
\begin{table*}[!t]
\caption{Cohort characteristics overall and by ESBL-phenotype status (n\,=\,132{,}955). ESBL+/ESBL$-$ denote presence/absence of the ESBL phenotype. Values are \% unless noted; SMD\,=\,standardized mean difference (SMD\,$>$\,0.1 = meaningful imbalance). ADI is the median among geocoded cultures.}
\label{tab:cohort}
\begin{center}
\footnotesize
\begin{tabular}{@{}lcccc@{\hspace{1.4em}}lcccc@{}}
\toprule
\textbf{Characteristic} & \textbf{All} & \textbf{ESBL+} & \textbf{ESBL$-$} & \textbf{SMD} &
\textbf{Characteristic} & \textbf{All} & \textbf{ESBL+} & \textbf{ESBL$-$} & \textbf{SMD} \\
\midrule
N (cultures)         & 132{,}955 & 19{,}159 & 113{,}796 & --    & Prior ESBL              & 16.6 & 40.4 & 12.5 & +0.67 \\
ESBL prevalence      & 14.41 & 100 & 0 & --                     & Prior CRE               & 3.1  & 6.4  & 2.5  & +0.19 \\
Age $\geq$65         & 56.2 & 59.0 & 55.7 & +0.07                & Prior MRSA              & 3.2  & 6.1  & 2.7  & +0.17 \\
Male sex             & 24.2 & 36.2 & 22.2 & +0.31                & Prior VRE               & 2.4  & 5.9  & 1.8  & +0.21 \\
Inpatient            & 31.0 & 47.7 & 28.1 & +0.41                & No. prior orgs (median) & 1    & 1    & 1    & +0.31 \\
Outpatient           & 46.0 & 32.9 & 48.2 & $-$0.32             & Fluoroquinolone (90d)   & 11.8 & 20.9 & 10.2 & +0.30 \\
Emergency dept       & 17.0 & 13.1 & 17.6 & $-$0.13             & 3rd-gen ceph. (90d)     & 28.0 & 43.8 & 25.4 & +0.39 \\
Urgent care          & 5.7  & 5.9  & 5.7  & +0.01                & Carbapenem (90d)        & 4.6  & 16.3 & 2.6  & +0.48 \\
ADI (geocoded, med.) & 17.8 & 17.3 & 17.8 & +0.01                & Glycopeptide (90d)      & 13.6 & 28.8 & 11.1 & +0.46 \\
ADI ungeocoded       & 42.6 & 40.1 & 43.0 & $-$0.06             & Sulfonamide (90d)       & 8.8  & 14.1 & 7.9  & +0.20 \\
Elixhauser (median)  & 0    & 0    & 0    & +0.01                & Ext.-spectrum PCN (90d) & 11.6 & 20.6 & 10.1 & +0.30 \\
Renal failure        & 8.5  & 8.5  & 8.5  & 0.00                 & Aminoglycoside (90d)    & 1.1  & 2.1  & 1.0  & +0.09 \\
\bottomrule
\end{tabular}
\end{center}
\end{table*}

\subsection*{Model Performance}
Cost-sensitive learning outperformed SMOTE across all model families (XGBoost: AUROC 0.760 vs.\ 0.725), but it did not outperform no correction at all: an otherwise identical XGBoost without class weighting reached AUROC 0.762 (patient-clustered difference versus the cost-sensitive model $-$0.002, 95\% CI $-$0.005 to $-$0.0002) with raw probabilities that were already calibrated (mean predicted 0.141 vs.\ observed 0.137; Brier 0.099). Class weighting therefore bought no discrimination and only inflated the probability scale, consistent with simulation evidence~\cite{goorbergh2022}. All subsequent results use the cost-sensitive models as originally specified, with probabilities recalibrated as described. XGBoost outperformed both comparator models across all metrics (Table~\ref{tab:performance}). On the patient-disjoint held-out test set (n\,=\,26,369 cultures from 14,475 patients), XGBoost achieved AUROC 0.760 (95\% CI: 0.747--0.772) and AUPRC 0.430 (95\% CI: 0.399--0.459). Logistic regression and random forest achieved AUROC 0.730 (95\% CI: 0.717--0.743) and 0.740 (95\% CI: 0.726--0.754), respectively. Patient-grouped cross-validation confirmed stability (AUROC 0.759\,$\pm$\,0.008 across five folds). As a sensitivity analysis, the culture-level random split inflated every model: XGBoost rose to AUROC 0.773 / AUPRC 0.472 and the tuned model to 0.773 / 0.465. Within it, discrimination was far higher for test cultures whose patient appeared in training than for unseen patients (0.809 vs.\ 0.678), though case mix also differs (prevalence 16.9\% vs.\ 10.7\%); we therefore report patient-grouped estimates throughout.

\begin{table}[!t]
\caption{Performance on the patient-disjoint test set (26,369 cultures, 14,475 patients; prevalence 13.74\%). Cost-sensitive learning; patient-clustered bootstrap CIs; Brier for raw scores (tuned model recalibrates to 0.098).}
\label{tab:performance}
\begin{center}
\begin{tabular}{|l|c|c|c|}
\hline
\textbf{Model} & \textbf{AUROC (95\% CI)} & \textbf{AUPRC (95\% CI)} &
\textbf{Brier} \\
\hline
Logistic Regression & 0.730 (0.717--0.743) & 0.383 (0.354--0.413) & 0.197 \\
\hline
Random Forest       & 0.740 (0.726--0.754) & 0.403 (0.373--0.434) & 0.194 \\
\hline
XGBoost             & 0.760 (0.747--0.772) & 0.430 (0.399--0.459) & 0.175 \\
\hline
XGBoost (tuned)     & \textbf{0.763} (0.750--0.775) & \textbf{0.435} (0.405--0.465) & 0.182 \\
\hline
\end{tabular}
\end{center}
\end{table}

\begin{figure}[!t]
\centering
\includegraphics[width=0.60\columnwidth]{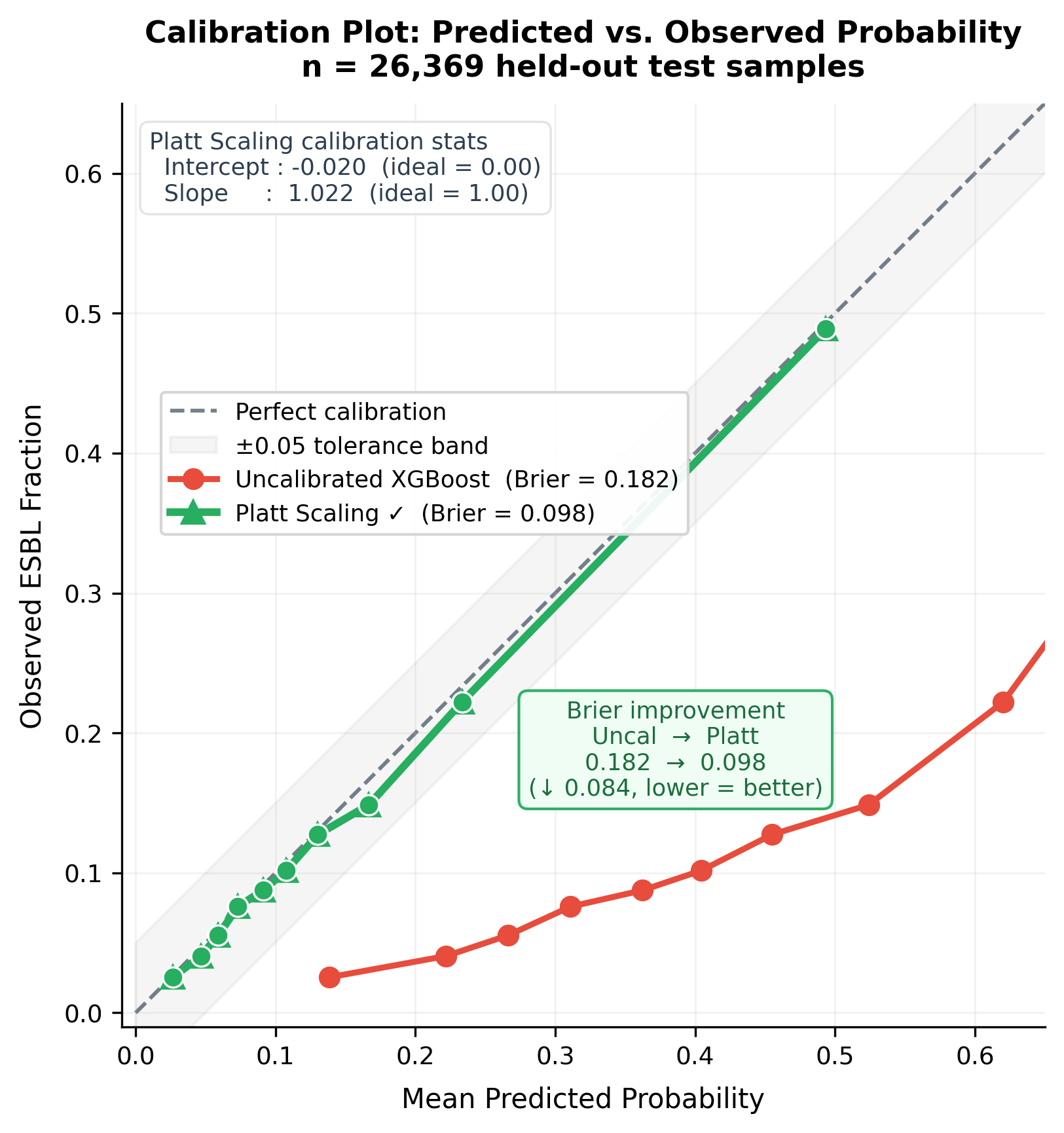}
\caption{Calibration reliability diagram (tuned XGBoost, n\,=\,26,369). Platt-recalibrated probabilities (green) track the diagonal; raw cost-weighted scores (red) overestimate risk throughout.}
\label{fig:calibration}
\end{figure}

\noindent\textbf{Hyperparameter optimization.}~Optuna tuning with patient-grouped inner folds improved XGBoost AUROC from 0.760 (default) to 0.763 ($\Delta$\,=\,+0.003). The tuned model was adopted as the final model. Raw cost-weighted scores over-predicted risk (mean predicted 0.41 vs.\ observed 0.14; calibration intercept $-$1.77, slope 1.07). Platt recalibration reduced the tuned model's Brier score from 0.182 to 0.098 (null model 0.119), with post-recalibration intercept $-$0.02 and slope 1.02; isotonic regression was equivalent (Brier 0.098, intercept $-$0.03, slope 1.01) (Figure~\ref{fig:calibration}).

\vspace{-4pt}
\vspace{-8pt}

\subsection*{Clinical Operating Points}
At the \textit{alert} threshold (calibrated p\,$\geq$\,0.068), the model achieved 90.1\% sensitivity, 35.6\% specificity, and \textbf{95.8\% NPV}, flagging 17,902 of 26,369 test-set cultures with 359 phenotype-positive cultures missed. At the \textit{balanced} threshold (p\,$\geq$\,0.227, maximizing F$_1$), sensitivity was 45.6\%, specificity 89.4\%, PPV 40.8\%, and NPV 91.2\%. The 95.8\% NPV at the alert threshold implies that a screen-negative result carries a post-test ESBL probability of 4.2\%, which may support empiric narrow-spectrum therapy in non-ICU settings where the baseline ESBL risk is below the carbapenem initiation threshold \cite{tamma2024}.

\begin{table}[!t]
\caption{Clinical operating points for tuned XGBoost on the patient-disjoint test set (n\,=\,26,369). Cutoffs are on the Platt-recalibrated probability scale; the alert threshold mirrors the $\geq$90\% susceptibility standard from institutional antibiograms.}
\label{tab:operating_points}
\begin{center}
\small
\begin{tabular}{@{}lccccc@{}}
\toprule
\textbf{Threshold} & \textbf{Calibrated p} & \textbf{Sens (\%)}  & \textbf{Spec (\%)} & \textbf{PPV (\%)} & \textbf{NPV (\%)} \\
\midrule
Alert ($\geq$90\% Sens)       & 0.068 & 90.1 & 35.6 & 18.2 & \textbf{95.8} \\
Balanced (max F$_1$)           & 0.227 & 45.6 & 89.4 & 40.8 & 91.2 \\
Conservative ($\geq$80\% Spec) & 0.165 & 57.0 & 80.0 & 31.2 & 92.1 \\
\bottomrule
\end{tabular}
\end{center}
\end{table}

\noindent\textbf{Order-time partition.}~Partitioning by culture order time gave XGBoost AUROC 0.766 (LR 0.734, RF 0.745), matching the patient-grouped estimates. This is not temporal validation: ARMD-MGB shifts dates per patient (177-year apparent span for a 10-year collection; median within-patient span 0.00 years), so the ordering reflects random offsets; its effect is to make the partition 97.5\% patient-disjoint, which is why it agrees with the grouped rather than the inflated result.

\noindent\textbf{Organism-stratified performance.}~The pooled model maintained discrimination across organism strata. AUROC was 0.763 for \textit{E.\ coli}, 0.784 for \textit{K.\ pneumoniae}, and 0.710 for other Enterobacteriaceae.

\noindent\textbf{Decision curve analysis.}~On recalibrated probabilities, XGBoost's net benefit exceeded treat-all, treat-none, logistic regression and random forest at every threshold probability from 0.01 to 0.80; at a threshold of 0.10 net benefit was 0.065 versus 0.042 for treat-all, and at 0.15 it was 0.048 versus $-$0.015. At the alert threshold (calibrated p\,$\geq$\,0.068) the model withheld broad-spectrum therapy in 321 of every 1,000 cultures, 307 of them phenotype-negative and 14 phenotype-positive, a net reduction of 121 unnecessary courses per 1,000 relative to treat-all at that threshold~\cite{vickers2006}.

\noindent\textbf{Statistical comparison.}~DeLong tests confirmed that XGBoost significantly outperformed both logistic regression (z\,=\,13.87, p\,$<$\,10$^{-43}$) and random forest (z\,=\,12.71, p\,$<$\,10$^{-36}$). Because DeLong's test assumes independent observations, we also computed patient-clustered bootstrap intervals for the AUROC differences: tuned XGBoost exceeded logistic regression by 0.033 (95\% CI 0.027--0.039) and random forest by 0.023 (0.018--0.028), and the gain from tuning over default XGBoost was 0.003 (0.000--0.006), with 2.2\% of resamples showing no gain. On recalibrated probabilities, category-based NRI for XGBoost versus logistic regression was +0.112 (95\% CI: +0.093 to +0.131), driven by events reclassified upward (event NRI +0.150) with a small loss among non-events ($-$0.038); continuous NRI was +0.036 (95\% CI: +0.033 to +0.039).

\subsection*{Feature Importance}
SHAP analysis on the tuned model revealed prior ESBL colonization as the dominant predictor (mean $|$SHAP$|$\,=\,0.389), followed by number of prior cultured organisms (0.197), ADI (0.191), multi-drug-resistant organism history (0.185), and days since prior culture (0.155). Established clinical risk factors, including prior carbapenem (0.074) and cephalosporin (0.069) exposure, ranked substantially lower. Notably, ADI alone discriminated at chance on geocoded cultures (AUROC 0.496) despite its third-place SHAP rank, indicating its contribution is interaction-driven, adding signal jointly with clinical features rather than as a standalone predictor. ADI ranked second under the culture-level split; patient-level partitioning moves prior organism burden ahead of it, consistent with part of its apparent importance reflecting within-patient stability.

\vspace{-4pt}
\begin{figure}[!t]
\centering
\includegraphics[width=0.75\columnwidth]{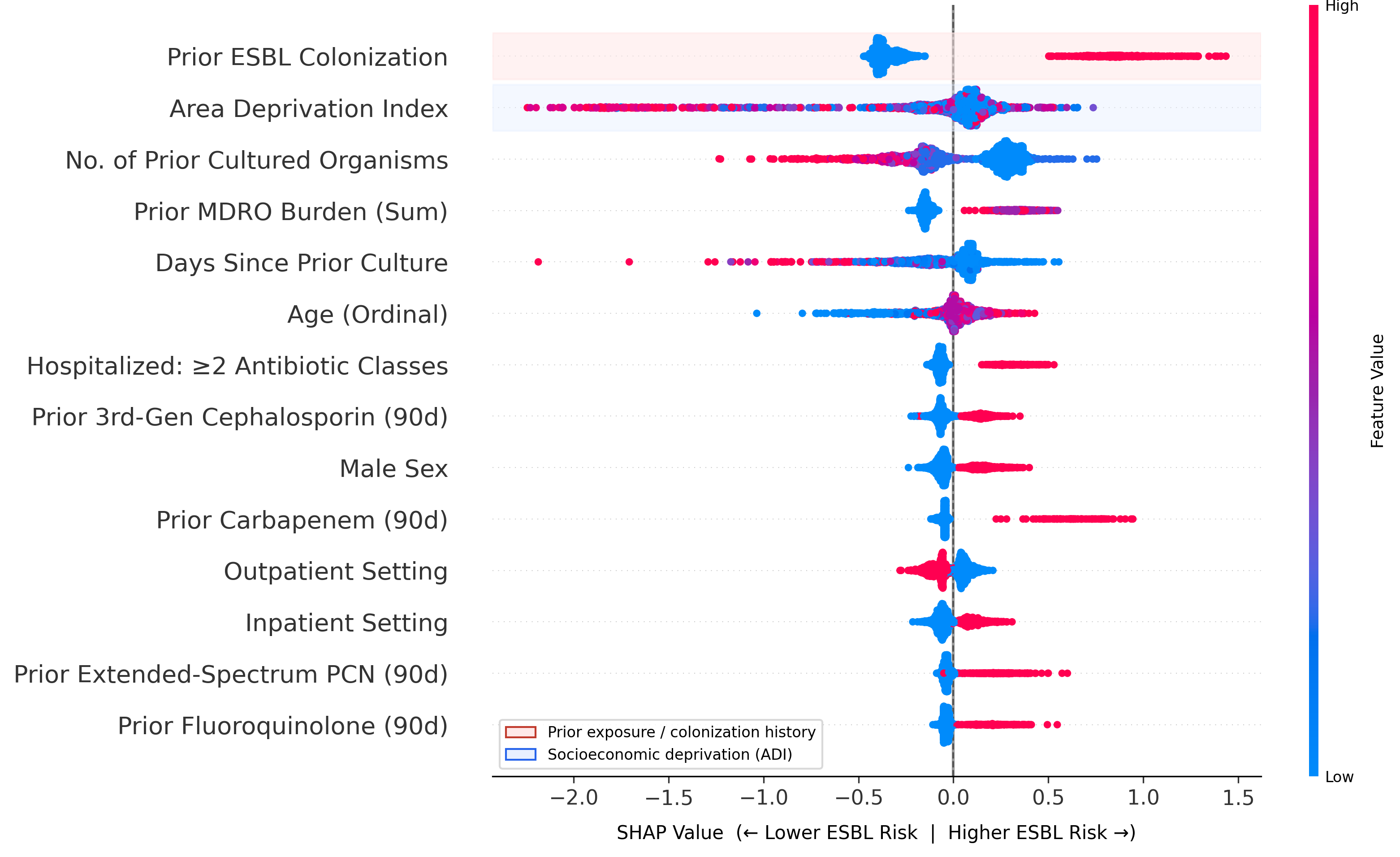}
\caption{SHAP beeswarm plot for the tuned XGBoost model 
(n\,=\,2,000 held-out test samples). Each point represents 
one patient's feature contribution to ESBL risk prediction. 
Prior ESBL colonization is the dominant predictor, followed 
by prior organism burden and Area Deprivation Index, with all 
remaining features ranking substantially lower.}
\label{fig:shap_beeswarm}
\end{figure}
\vspace{-6pt}

\noindent\textbf{Subgroup performance.}~Model performance varied across clinical subgroups. Performance was highest in patients with prior CRE (AUROC 0.814), prior ESBL (AUROC 0.804), and those in the most deprived ADI quintile (AUROC 0.820). Performance was attenuated in urgent care settings (AUROC 0.689) and outpatient cultures (AUROC 0.734), likely reflecting sparser clinical histories. A sex-based performance gap persisted (male AUROC 0.766 vs.\ female AUROC 0.746; $\Delta$\,=\,0.020).

\noindent\textbf{ESBL-naive cohort.}~On the ESBL-naive sub-cohort (no prior ESBL history; 22,178 test-set cultures, 84.1\% of test set; ESBL prevalence 10.14\%), XGBoost (tuned) achieved AUROC 0.696 (95\% CI: 0.684--0.707; AUPRC 0.220, 95\% CI: 0.206--0.237), a decrease of $\Delta$AUROC\,=\,$-$0.067 versus the full cohort, confirming meaningful risk signals from antibiotic exposure, comorbidity, and neighborhood deprivation independent of prior colonization.

\noindent\textbf{Ungeocoded addresses and ADI missingness.}~ADI was missing for 42.6\% of cultures, and missingness was non-random: ungeocoded cultures were less often inpatient (16.6\% vs.\ 41.6\%; SMD\,0.57), more often outpatient (56.8\% vs.\ 38.0\%) or urgent care, younger, and less likely to have prior ESBL (11.6\% vs.\ 20.2\%; SMD\,0.24). Discrimination was correspondingly higher in geocoded (AUROC 0.788) than ungeocoded cultures (0.724), reflecting richer documented histories rather than deprivation itself; the missingness indicator ranked 14/45 by gain, and the 42-feature no-ADI model removes the geocoding dependency entirely.

\noindent\textbf{ADI ablation and fairness.}~Removing all three ADI features reduced XGBoost (tuned) from AUROC 0.763 to 0.743 (95\% CI: 0.733--0.751; $\Delta$AUROC\,=\,$-$0.020; $\Delta$AUPRC\,=\,$-$0.018). The modest decrease indicates ADI's signal is largely recaptured by correlated clinical features. Combined with 42.6\% missing ADI, these results support deploying a 42-feature model without ADI to avoid encoding socioeconomic bias into treatment decisions~\cite{banerjee2023shortcuts}.

\noindent\textbf{Enzyme-class sensitivity analysis.}~Under the enzyme-class definition (\texttt{enzyme\_class}~=~``ESBL''), prevalence was 6.09\% (n\,=\,8,091) versus 14.41\% for the ESBL phenotype. XGBoost achieved AUROC 0.879, consistent with the narrower case definition, making classification easier. This confirms that ESBL definition choice materially affects both prevalence and apparent performance, supporting the phenotype definition as primary, given its clinical relevance.

\noindent\textbf{Failure analysis.}~At the alert threshold, the 359 missed phenotype-positive cultures were not the clinically complex cases. Compared with the 3,263 caught, they had far less documented history (prior ESBL 1.1\% vs.\ 42.0\%; prior carbapenem 0.0\% vs.\ 17.3\%; multi-resistant history 3.6\% vs.\ 60.6\%; third-generation cephalosporin exposure 13.7\% vs.\ 47.5\%; all p\,$<$\,10$^{-16}$), were more often outpatient (63.5\% vs.\ 30.1\%) and had a longer interval since any prior culture (mean 245 vs.\ 81 days), whereas comorbidity burden and prior organism counts did not differ (p\,$>$\,0.2). The model fails on history-poor cultures, not on heavily pre-treated patients.

\subsection*{Additional Analyses}
\noindent\textbf{ADI encoding.}~Refitting the tuned model with ADI as a native missing value instead of zero left discrimination unchanged (AUROC 0.763, AUPRC 0.437) and moved ADI from third to second in mean $|$SHAP$|$ (0.210 vs.\ 0.191). In both encodings the ADI attribution sits on geocoded cultures (mean $|$SHAP$|$ 0.27--0.29 on geocoded vs.\ 0.08--0.10 on ungeocoded rows), so the zero-fill convention is not the source of ADI's importance. Ablating the ADI score and high-deprivation flag while retaining the missingness indicator cost $\Delta$AUROC $-$0.015 (0.763 to 0.748); ablating all three ADI features cost $-$0.020. ADI's contribution is therefore real but conditional: it has no marginal association with the outcome (AUROC 0.496 alone; SMD 0.005), yet it improves discrimination jointly with clinical features, a pattern consistent with interaction effects or with ADI acting as a proxy for unmeasured geography such as the contributing hospital.

\noindent\textbf{Specimen type.}~Urine accounted for 89.1\% of cultures, blood for 6.5\% and respiratory specimens for 4.4\%. The reported model discriminated better in the higher-prevalence specimens: AUROC 0.745 in urine (prevalence 12.2\%), 0.789 in blood (23.3\%; AUPRC 0.613) and 0.792 in respiratory cultures (30.8\%; AUPRC 0.694). Adding three specimen indicators to the feature set did not improve the pooled model (AUROC 0.764 vs.\ 0.763; patient-clustered difference +0.001, 95\% CI $-$0.000 to +0.002), and the indicators ranked 22nd to 27th of 48 features, indicating that the care-setting features already carry this information.

\noindent\textbf{Strict ESBL-E definition.}~The four IDSA ESBL-E species accounted for 118,411 cultures (89.1\%) and 16,176 of the 19,159 phenotype-positive cultures (84.4\%); the AmpC-type genera contributed 2,872 positives (15.0\%) at a within-genus prevalence of 20.8\%. Ceftriaxone was tested in 84.9\% of IDSA-species cultures; among them every ceftriaxone-resistant culture was phenotype-positive, while 2,513 phenotype-positive cultures were ceftriaxone-susceptible (composite prevalence 16.1\% vs.\ 13.6\% for ceftriaxone resistance). Performance was insensitive to the definition: the reported model achieved AUROC 0.769 on IDSA species with the composite label, 0.700 on the remaining genera, and 0.764 when scored against ceftriaxone resistance alone, and a model retrained on the strict label reached 0.766 (95\% CI 0.751--0.779; AUPRC 0.438, 95\% CI 0.402--0.473) against 0.763 for the primary analysis.

\section*{Discussion}
We developed and validated an XGBoost model predicting an ESBL phenotype across 132,955 Enterobacteriaceae cultures from 12 hospitals, achieving AUROC 0.763 with 95.8\% NPV at the 90\%-sensitivity threshold. Larger multi-site work exists for per-agent susceptibility prediction~\cite{goto2026,yelin2019}; to our knowledge, this is the largest multi-hospital study focused on an ESBL-phenotype outcome that reports patient-level partitioning and quantifies neighborhood deprivation against clinical history.

\paragraph{Comparison with prior work.}
Our AUROC of 0.763 is consistent with recent EHR-based models of extended-spectrum cephalosporin resistance: 0.74 for ceftriaxone resistance in Oxford bacteraemia cohorts~\cite{yuan2025}, 0.78--0.81 in the Veterans Health Administration multitask model, which also used patient-level and prospective partitions~\cite{goto2026}, and 0.80 in an emergency-department ESBL model~\cite{kuzmich2025}. Our culture-level sensitivity split yields 0.773, so where prior studies split at the culture or isolate level their estimates are likely optimistic by a similar margin. The modest discrimination reflects the difficulty of predicting a phenotype with diverse mechanisms from pre-culture features alone.

\paragraph{Prior ESBL colonization as dominant predictor.}
Consistent with the primacy of personal culture history in prior work~\cite{yelin2019,macfadden2018}, and unlike ecological studies that foreground neighborhood deprivation~\cite{cooper2024,henderson2025}, our SHAP analysis identified prior ESBL colonization as the single most important predictor, with prior organism burden second and ADI third. Stewardship programs should therefore prioritize systematic documentation and retrieval of prior resistance history at culture ordering. Performance on the ESBL-naive sub-cohort (AUROC 0.696) confirms meaningful risk signals beyond prior colonization alone, where the clinical question is hardest.

\paragraph{ESBL phenotype vs.\ enzyme-class definition.}
The narrower enzyme-class definition yields higher apparent discrimination (Results), but part of that gain is likely ascertainment: confirmatory enzyme testing is more often performed in patients with a known resistance history, which is exactly what the model predicts well. We advocate the phenotype definition as primary because treatment decisions are based on the susceptibility phenotype, susceptibility data are universally available, and this definition aligns with prior validation studies~\cite{rottier2018,yuan2025}.

\paragraph{Clinical utility.}
DCA on recalibrated probabilities showed net benefit over treat-all and treat-none throughout the examined range, with the largest gains between 0.05 and 0.20, where empiric carbapenem decisions are typically made. The 95.8\% NPV reflects a higher baseline prevalence than enzyme-class models and may be acceptable in non-ICU settings. This 90\%-sensitivity operating point mirrors the $\geq$90\% susceptibility standard used by institutional antibiograms, grounding the threshold in established practice.

\paragraph{Clinical deployment and Algorithmic Equity.}
We envision two EHR integrations: an active Best Practice Advisory at order entry for high-acuity settings, and a passive stewardship dashboard for pharmacist review to limit alert fatigue; both require prior ESBL history to be computable at $t_0$. Because an AUROC of 0.763 supports rather than replaces clinical judgment, a screen-negative result should guide de-escalation only in non-ICU settings and must never withhold escalation in ICU, sepsis, or suspected bloodstream infection. False negatives are not the complex, heavily pre-treated patients, who are flagged, but history-poor, mostly outpatient cultures without documented colonization or recent broad-spectrum exposure; a screen-negative result cannot lower risk below what the structured record supports, so local prevalence and unrecorded exposures such as travel or care elsewhere still matter. Local re-validation and recalibration are mandatory before deployment at any new site, with periodic recalibration as resistance epidemiology drifts.

ADI plays a dual role. Epidemiologically, its third-place SHAP rank (0.191) implicates neighborhood-level determinants, healthcare access, environmental exposure, and crowding, in ESBL risk, or unmeasured geography such as the contributing hospital. For deployment, however, conditioning empiric therapy on where a patient lives risks encoding algorithmic bias. This tension is resolvable: removing all three ADI features costs only $\Delta$AUROC\,$-$0.020, since the signal is largely recaptured by correlated clinical features. We therefore report ADI as a modifiable community-level factor for population-health strategy while deploying the 42-feature no-ADI model at the bedside to preserve equity.

\paragraph{Robustness of the main findings.}
Four checks strengthen the main findings. Class weighting was unnecessary: an uncorrected XGBoost matched the cost-sensitive model's discrimination with calibrated raw probabilities, so the recalibration step repairs a distortion that the imbalance correction itself introduced~\cite{goorbergh2022}, and future versions can drop the weighting. The composite ESBL phenotype is not driving the results, since the strict IDSA definition gave the same AUROC. Specimen type adds no information beyond care setting, although the model is most accurate where the carbapenem decision matters most, in blood and respiratory cultures. Finally, ADI's importance survives a change of missing-value encoding, which rules out a zero-fill artefact but not a site-proxy explanation.

\paragraph{Limitations.}
First, the 12-hospital MGB network is one New England system, and because the release lacks a site identifier, cross-hospital case mix could not be adjusted for; robustness rests on organism-stratified (0.71--0.78) and subgroup stability pending external validation. Second, ADI was missing non-randomly for 42.6\% of cultures (ungeocoded addresses skewed lower-acuity and outpatient), and discrimination was lower in this subset (0.724 vs.\ 0.788); the deployable no-ADI model removes this geocoding dependency entirely. Third, the primary outcome is a composite ESBL phenotype that includes piperacillin-tazobactam resistance and AmpC-producing genera; the strict ESBL-E sensitivity analysis shows that this choice does not change discrimination (0.766 vs.\ 0.763), but the composite label must not be read as a molecular ESBL diagnosis. Fourth, because ARMD-MGB shifts dates per patient, we cannot assess calendar-time drift; the model's stability as resistance epidemiology evolves is therefore untested and must be established prospectively or in an external cohort with unshifted timestamps.

\paragraph{Future directions.}
Prospective validation in a clinical decision support system is the critical next step, followed by external validation in other health systems. Specimen type, prior susceptibility results, unstructured notes, and travel or prior-hospitalization history may further improve discrimination, and causal mediation analysis of the ADI-ESBL association could inform geographically targeted stewardship.

\begin{center}
{\normalsize\textbf{References}}
\end{center}

\begingroup
\renewcommand{\section}[2]{}  
\bibliographystyle{vancouver}
\bibliography{amia}
\endgroup

\end{document}